\pdfoutput=1

\documentclass[11pt]{article}

\usepackage[]{latex/acl}

\usepackage{times}
\usepackage{latexsym}

\usepackage{fontawesome}
\usepackage[T1]{fontenc}

\usepackage[utf8]{inputenc}

\usepackage{microtype}

\usepackage{inconsolata}

\usepackage{graphicx}

\usepackage{xspace}
\usepackage{tabularray}
\usepackage{amsmath}
\usepackage{algorithm}
\usepackage{algorithmic}
\usepackage{multirow}
\usepackage{paralist}
\usepackage{xcolor}
\usepackage{booktabs}
\usepackage{tabularx}
\usepackage{booktabs}

\newcommand{\remove}[1]{}

\usepackage{authblk}
\usepackage{times}  

\title{Visual Editing with LLM-based Tool Chaining:  \\ An Efficient Distillation Approach for Real-Time Applications}

\usepackage{authblk}

\author{\textbf{Oren Sultan}\thanks{Work  done during an internship at Lightricks.}$^{1,2}$, \textbf{Alex Khasin}$^{2}$, \textbf{Guy Shiran}$^{2}$
\\
\textbf{Asnat Greenstein-Messica}$^{2}$, \textbf{Dafna Shahaf}$^{1}$ \\
$^{1}$The Hebrew University of Jerusalem, $^{2}$Lightricks \\ 

\texttt{\{oren.sultan,dshahaf\}@cs.huji.ac.il} \\
\texttt{\{osultan,akhasin,gshiran,asi\}@lightricks.com} 
}

\begin{document}
\maketitle

\begin{abstract}

We present a practical distillation approach to fine-tune LLMs for invoking tools in real-time applications. We focus on visual editing tasks; specifically, we modify images and videos by interpreting user stylistic requests, specified in natural language (``golden hour''), using an LLM to select the appropriate tools and their parameters to achieve the desired visual effect.

We found that proprietary LLMs such as GPT-3.5-Turbo show potential in this task, but their high cost and latency make them unsuitable for real-time applications.
In our approach, we fine-tune a (smaller) student LLM with guidance from a (larger) teacher LLM and behavioral signals.
We introduce offline metrics to evaluate student LLMs. 
Both online and offline experiments show that our  student models succeeded in matching the performance of our teacher model (GPT-3.5-Turbo), significantly reducing costs and latency.
Lastly, we show that fine-tuning was improved by 25\% in low-data regimes  using augmentation.

\end{abstract}

\section{Introduction}

\begin{figure*}[t]
\includegraphics[width=.99\textwidth]{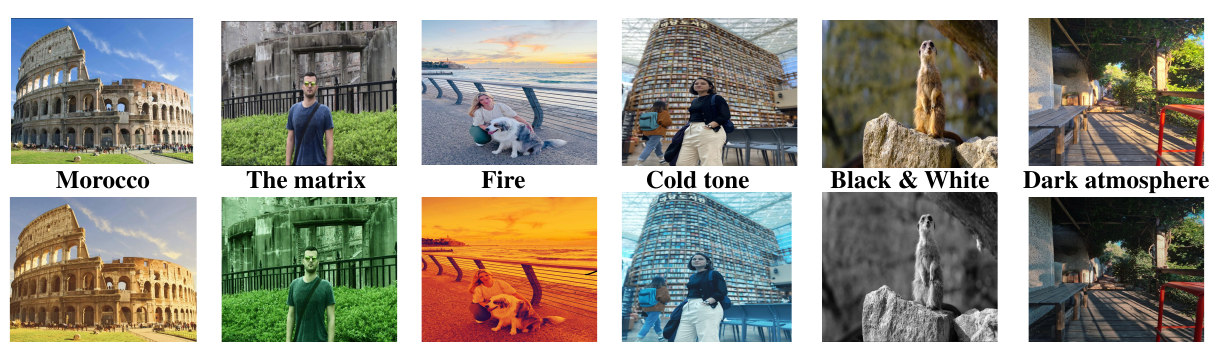}
\caption{
An illustration of our visual editing task. Users input an image/video and specify the desired visual appearance (\textbf{upper row: source images, middle: user intents}). An LLM interprets these intents, selects tools, and sets parameters. The \textbf{bottom} row displays the generated images by applying the LLM's output in our app.  For example, inputting ``Morocco'' \textbf{(left)} results in warm hues typical of Moroccan landscapes, reflecting its deserts.}
\label{fig:ai_recolor_examples}
\end{figure*}



Videos are a powerful communication and storytelling medium, gaining popularity through social media and video-sharing platforms. This surge has inspired many to create content. However, the complexity of video editing, with its numerous parameters and their interactions, poses significant barriers for beginners \cite{zhang2022ai}. 

Using natural language as an interaction medium for video editing can mitigate this challenge. Text-to-video, diffusion-based models that support instruction-guided video editing have demonstrated impressive results. However, they are computationally expensive, slow, and still lack in visual quality and user control over the generated video \cite{geyer2023tokenflow, couairon2023videdit, qi2023fatezero}. This makes them unsuitable for real-time mobile applications, which need to combine high editing quality, low execution cost and fast response.



We believe that instead of relying on an end-to-end approach that treats deep learning models as black boxes, it is more beneficial to teach LLMs to use \emph{existing, specialized} tools. This approach is also more \emph{interpretable}.
We are encouraged by recent advances in LLMs that demonstrated the effectiveness of building AI agents that leverage multiple external tools with LLMs \cite{schick2024toolformer,wang2023voyager,openai_assistant_website}, in particular for vision or vision-language  tasks \cite{liu2023llava,yang2024gpt4tools,wu2023visual}. 


In our work, we leverage LLMs to invoke existing, \emph{traditional} video editing tools that are \emph{specialized} for our task. Our aim is to implement an AI assistant in our video editing mobile app, democratizing advanced capabilities. As a proof-of-concept, we focused on \emph{tonal color adjustments}, allowing users to change a video's  appearance via textual instructions (e.g., ``golden hour''; see Figure~\ref{fig:ai_recolor_examples}).


Learning tool chaining through prompt engineering and in-context learning often relies on proprietary LLMs like GPT-3.5 \cite{yang2024gpt4tools}. These models are expensive, not publicly available, and slow, posing significant challenges for online production systems.
We propose a distillation approach based on fine-tuning an open source (smaller) student LLM for tools usage using the output from a (larger) teacher LLM, enhanced by user behavioral signals. 

We create offline metrics to evaluate model performance, involving the choice of tools to apply  and their parameters. This evaluation is challenging due to continuous parameter values and to our creativity-focused use case, with no single correct answer.
Finally, we develop a data augmentation scheme and demonstrate a 25\% improvement in the common real-life scenario of low-data regimes.
\noindent\textbf{Our contributions are:} (1)  We propose a practical distillation method to fine-tune open-source (smaller) student LLM for invoking tools, using a (larger) teacher LLM and behavioral feedback. We demonstrate the effectiveness of our approach in real-time production settings for visual editing. Our solution achieves low cost and latency, making it suitable for industry applications.
(2) We develop offline evaluation metrics for complex LLM tool chaining. 
(3)  Our experiments, both online and offline, 
show that our smaller student models succeeded in matching the performance of our teacher, GPT-3.5-Turbo. 
Additionally, we show a 25\% improvement in fine-tuning in low-data regimes using data augmentation.
(4) Our code and dataset are publicly available at our project website: \url{ https://www.orensultan.com/ai_recolor.github.io/}.


\section{Problem Statement}
\label{sec:problem_statement}

\begin{figure*}[t]
\includegraphics[width=0.9\textwidth]{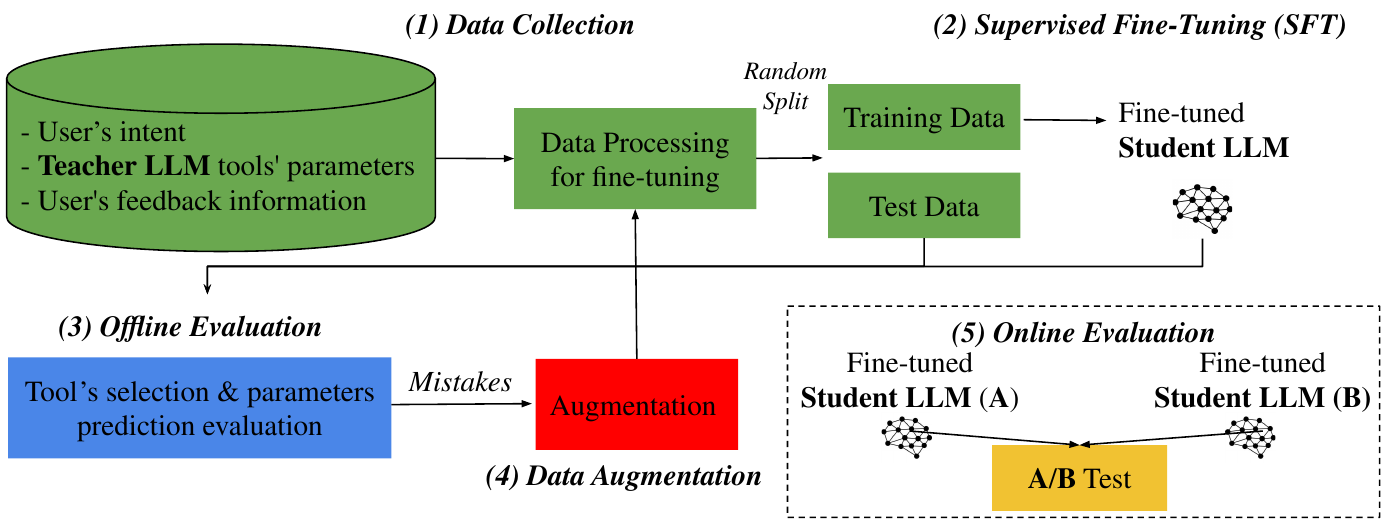}
\centering
\caption{Our distillation framework approach.
\textbf{(1)} We create a dataset by collecting user intents and the output (or potentially multiple outputs, if several users expressed the same intent) of our \emph{teacher LLM}.  We ensure high quality by keeping outputs users chose to export frequently (one output with the highest export rate per intent).  After data processing, we randomly split the data into \emph{training} and \emph{test} sets.
\textbf{(2)} We fine-tune a smaller \emph{student LLM} on our dataset.
\textbf{(3)} Offline, we evaluate the student LLM's selection of tools and predicted parameters.
\textbf{(4)} To improve fine-tuning in low-data regimes, we use an LLM to augment the training data by generating \emph{similar} samples (e.g., ``cool tone'' from ``cool morning'') to mistakes of the \emph{student LLM}.
\textbf{(5)} If a better \emph{student model} is found offline, we conduct an online A/B test.
}
\label{fig:ft_pipeline}
\end{figure*}

Our visual editing task deals with \emph{color grading} -- a post-processing procedure that alters the appearance of an image or a video by adjusting its tonal colors.
Our application features three tonal adjustment tools: \emph{global adjust} (global color range), \emph{selective adjust} (selective color ranges), and \emph{filters}. Each tool has up to a dozen parameters, which can be difficult for beginners to set correctly.

In our task, the user provides an asset (image/video) and a free-text description of the requested appearance. This raises the following challenges: (1) How to interpret the user's intent, which can be vague or require specific knowledge (e.g., given ``The Matrix'' request, it should recognize the distinctive imagery associated with the movie, characterized by a green tint, high contrast, and cyberpunk aesthetic). (2) How to decide which tools to use and with what parameters and values.
More formally, 
the AI Assistant's function, \(f : I \to O\), maps a user's intent (\(I\)) into a tailored configuration of tools and settings (\(O\)), interpreting and implementing the user's intent. 

The output is of the following form:
\begin{equation}
O = \left\{ (T_i, P_i) \mid T_i \in \mathcal{T}, P_i \in \mathcal{P}(T_i) \right\}
\label{eq:output_eq}
\end{equation}
where \(\mathcal{T}\) is the set of the available tools, and \(\mathcal{P}(T_i)\) is the power set of all possible parameter-value pairs for the tool \(T_i\), including the empty set \(\emptyset\) for when the tool is not used.
We denote \(P_i\) as the set of parameter-value pairs for the i-th tool:
\begin{equation}
P_i = \{(p_{i_1}, v_{i_1}), (p_{i_2}, v_{i_2}), \ldots, (p_{i_n}, v_{i_n})\}
\label{eq:parameters_values_eq}
\end{equation}
where \(p_{i_j}\), \(v_{i_j}\) are \(T_i\)'s j-th parameter and value.

Figure~\ref{fig:ai_recolor_examples} shows examples of various input images (top), with intents (middle), and  outputs (bottom).
See Appendix~\ref{subsec:appendix_tools_parameters} for details on tool parameters.

\section{Our Distillation Framework Approach}

Our goal is to automate our visual editing task using LLMs. In our proof-of-concept, we found that proprietary LLMs, like GPT-3.5-Turbo, can solve this task using preliminary prompts (based on an evaluation conducted by five experts from our team, who assessed the results across 20 different inputs). However, their high cost and latency make them unsuitable for real-time industry applications.

We employ a distillation framework approach (see Figure~\ref{fig:ft_pipeline}). We \textbf{generate data} by collecting outputs from a \emph{teacher LLM} based on user intents. The teacher LLM selects relevant tools and sets their parameters. If multiple users express the same intent, this could result in multiple outputs per intent. We ensure high-quality data by retaining the best results based on user feedback, filtering out those with no engagement. The retained data samples are injected into prompts for fine-tuning and are randomly split into training and test sets {(\S\ref{subsec:data_collection})}.
Then, we \textbf{fine-tune} a much smaller \emph{student LLM} on this dataset {(\S\ref{subsec:fine_tuning})}.
Since comparing between fine-tuned models in an online A/B test is costly and takes time, we design \textbf{offline evaluation} {(\S\ref{subsec:offline_evaluation})} 
metrics to predict the model's performance online.
To improve fine-tuning in low-data settings, we use \textbf{augmentation} by having another LLM generate similar samples for those the \emph{student LLM} got wrong during \emph{training}, then add these to the \emph{training set} {(\S\ref{subsec:data_augmentation})}.
Finally, to compare the \emph{actual} performance of two fine-tuned student LLMs, we conduct an \textbf{online A/B test}  {(\S\ref{subsec:online_evaluation})}.

\subsection{Data Collection}
\label{subsec:data_collection}

Our goal is to collect high-quality data using a \emph{teacher LLM's} outputs to existing user intents for fine-tuning a \emph{student LLM}. \\
\noindent\textbf{Gathering Teacher LLM Outputs.}
We use GPT-3.5-Turbo\footnote{\url{https://platform.openai.com/docs/models/gpt-3-5-turbo}} as the teacher due to its cost/performance tradeoff. Initially, it was deployed in our video app,  serving users for four months, during which we collected data for fine-tuning.
A data row includes: (1) The user's intent with the requested vibe (e.g., ``x-ray''). (2) The output of the \emph{teacher LLM} to this intent, including the tools to use and their parameters. (3) Whether the user exports the result per tool (highly satisfied users export results). We filter out samples with zero exports ($\sim$80\%) to train on high-quality data. Our \emph{teacher LLM} can generate different outputs per intent (across different calls); we take as ground truth the result that maximizes the export rate.

In our teacher prompts, we included one-shot example for user intent, with an output of the rationale (a free text explanation of the reasoning, how to achieve the intent by adjusting parameters) as well as the output parameters for the tool. Integrating similar Chain-of-Thought (CoT) mechanisms has been shown to enhance LLMs' performance \cite{wei2022chain} and interpretability.
Refer to Appendix~\ref{subsec:appendix_teacher_llm_model} for our \emph{teacher LLM} implementation details, and Appendix~\ref{subsec:appendix_teacher_llm_prompts} Figures~\ref{tab:teacher_llm_adjust_prompt}, \ref{tab:teacher_llm_selective_adjust_prompt}, and \ref{tab:teacher_llm_filter_prompt} for the three prompts (one per tool) we used.

In total, we collected 9,252 unique user intents, each paired with corresponding {teacher} outputs for the three tonal adjustment tools, resulting in 27,756 data points. See Appendix~\ref{subsec:appendix_tools_parameters_values_distribution} for statistics on the distribution of different parameter's values across the different tools observed in the dataset.

\noindent\textbf{Data Processing for Fine-Tuning.}
We used the collected data to fine-tune a \emph{student LLM}, using three prompts. These prompts for the student were more concise those for the teacher, as the student would be fine-tuned on thousands of examples (instead of one-shot). We decided not to request rationale from the student, as we prioritize low latency, and generating the reasoning significantly increases the response time.
See Appendix~\ref{subsec:appendix_student_llm_prompts} Figures~\ref{tab:student_llm_adjust_prompt}, \ref{tab:student_llm_selective_adjust_prompt}, and \ref{tab:student_llm_filter_prompt} for the three prompts (one per tool) we used for a Llama-2-7b-chat-hf \emph{student LLM}.  

Note that the \emph{student LLM} is trained on all three tools  (similar to multi-task instruction), resulting in a unified model capable of predicting all tools.

\noindent\textbf{Data Splitting.}
We randomly split the data for fine-tuning into two disjoint sets: a \emph{test set} with 1K \emph{unique} user intents, each with a corresponding \emph{teacher LLM} output for each tool (3K samples), and a \emph{training set} with the remaining data, 8,252 rows.
Each row includes a user intent and three tool outputs.
%
Table~\ref{tab:dataset_distribution} 
shows the distribution of times each tool was used by the teacher.

\begin{table}[!t]
    \centering
    \footnotesize 
    \begin{tabular}{l cc cc cc}
        \toprule
        \textbf{Set} & \multicolumn{2}{c}{\textbf{Adjust}} & \multicolumn{2}{c}{\textbf{SelectiveAdjust}} & \multicolumn{2}{c}{\textbf{Filter}} \\
        & Used & All & Used & All & Used & All \\
        \midrule
        Train & 7570 & 8252 & 2647 & 8252 & 5448 & 8252 \\
        \midrule
        Test & 912 & 1000 & 356 & 1000 & 683 & 1000 \\
        \bottomrule
    \end{tabular}
    \caption{The train set has 8,252 rows of unique user intents, and the test has 1K. Each row includes a user intent and three tool outputs. ``Used'' indicates the number of rows where each tool should have been used.}
    \label{tab:dataset_distribution}
\end{table}



\subsection{Supervised Fine-Tuning (SFT)}
\label{subsec:fine_tuning}

Our goal is to fine-tune a \emph{student LLM} to mimic a \emph{teacher LLM} outputs (filtered using behavioral signals). Using our collected dataset for fine-tuning, \( D = \{(x, y)\} \), where \( x \) is the prompt (user's intent and task instructions) and \( y \) is the \emph{teacher LLM}'s output.  We fine-tune two types of LLMs.

\noindent\textbf{Auto-Regressive Model.}
We fine-tune a \emph{decoder-only} LLM to generate \( y = \{y_1, \ldots, y_n\} \)
using the \emph{auto-regressive} LLM objective, which maximizes the expected log-likelihood \cite{Radford2019LanguageMA}:
\begin{equation}
\mathcal{L}(\theta) = \sum_{t=1}^{T} \log P(y_t \mid y_{1}, y_{2}, \ldots, y_{t-1}; \theta)
\label{eq:sft_decoder_loss_eq}
\end{equation}
We aim to maximize the log probability of the target word \( y_t \) given prior words \((y_1, \ldots, y_{t-1})\) with model parameters \(\theta\). We used the Llama-2-7b-chat-hf \cite{touvron2023llama2} (see \ref{subsec:models} for details).

\noindent\textbf{Sequence-to-Sequence Model.}
We fine-tune an \emph{encoder-decoder} LLM to generate \( y = \{y_1, \ldots, y_n\} \)
using the \emph{sequence-to-sequence} LLM objective, which maximizes the expected log-likelihood  \cite{sutskever2014sequence}:
\begin{equation}
\mathcal{L}(\theta) = \sum_{t=1}^{T} \log P(y_t \mid y_{1}, y_{2}, \ldots, y_{t-1}, \mathbf{x}; \theta)
\label{eq:sft_encoder_decoder_loss_eq}
\end{equation}
We want to maximize the log probability of the target word \( y_t \) given the previous target words \((y_1, \ldots, y_{t-1})\) and the source sequence \(\mathbf{x}\), using model parameters \(\theta\).
We explored various sizes of FlanT5 \cite{chung2022scaling} aiming to keep high-quality results and reducing latency and GPU costs.

\subsection{Offline Evaluation}
\label{subsec:offline_evaluation}

Our goal is to evaluate the \emph{student LLM's} performance on our \emph{test set}. Since online evaluation (A/B testing) is time-consuming and costly, we design offline metrics to compare different \emph{student LLMs} and predict their performance in online A/B tests.

Our metrics assess two key elements of the task:
(1) \emph{Tool-selection:} the model's ability to decide correctly whether to use a tool. We measure precision and recall, and report the \emph{tool-selection score} as the F1-score. (2) \emph{Quality:} the model's ability to use a tool correctly. For the \emph{filter} tool, the \emph{quality score} is the \emph{accuracy} (proportion of correct predictions between the predicted and ground truth filter names).
For the \emph{adjust} and \emph{selective adjust} tools, the \emph{quality score} is the \emph{mean cosine similarity} across samples, on predicted and ground truth parameter values (where both prediction and ground truth agree the tool should be used). Note that this metric is overly strict, as a desired result might be achievable with different parameter combinations.

The \emph{final score} for a tool is the harmonic mean of the \emph{tool-selection score} and \emph{quality score}, emphasizing high performance in both. The \emph{overall score} is the average of the final scores of all tools.

For a reality check, we also analyze the actual generated images/videos by applying the tools' predicted parameters in our app. In this study, we analyze a random sample, with three human annotators per sample (see Section~\ref{subsec:results}, RQ1). Our ideas for automatic image evaluation, comparing two student LLMs, are provided in Appendix~\ref{subsec:appendix_gpt4v_image_evaluator}.

\subsection{Data Augmentation}
\label{subsec:data_augmentation}

A common industry need is fine-tuning a model with limited data. Here we demonstrate efficient data augmentation to improve this process.

Inspired by \citet{lee2024llm2llm}, we iteratively run the offline evaluation on the LLM's \emph{training set}. Each iteration involves two steps: (1) Identifying where the student LLM's predictions differ from the teacher's. For the \emph{filter} tool, a mistake occurs when the predicted filter name is incorrect. We define a mistake in the \emph{adjust} or \emph{selective adjust} tool when a sample's \emph{cosine similarity} is lower than the tool's \emph{mean cosine similarity} without data augmentation.
(2) Using another LLM to generate similar input user intents where the student LLM made mistakes (e.g., ``cool tone'' from ``cool morning''). These new intents, along with the teacher LLM's original answers, are added to the training set.

We evaluated the augmentation on different sizes of our training set (using random sampling).
To ensure a similar number of augmentations between different subsets of the training set, we always evaluated mistakes on a random sample of 1K. We augmented an intent if a mistake was identified in \emph{at least} one tool. Using GPT-4 \cite{openai2023gpt}, we generated similar user intents. Our implementation showed a 25\% performance improvement in low data regimes with just one iteration (Section~\ref{subsec:results}). See Appendix~\ref{subsec:appendix_aug_llm_model} 
for implementation details and Appendix~\ref{subsec:appendix_aug_llm_prompt}, Figure~\ref{tab:aug_llm_prompt} for the prompt used.

\subsection{Online Evaluation}
\label{subsec:online_evaluation}

When our \emph{offline evaluation} shows it is worthwhile to consider a new \emph{student LLM}, we recommend confirming this in an online A/B test experiment.

Our primary metric of interest is the \emph{project completion rate}, calculated as the number of \emph{projects exported} divided by the number of \emph{projects started}. This metric indicates total user satisfaction with the results and the overall experience.


\section{Experiments}

\label{subsec:research_questions}
We focus on the following research questions:

\noindent\textbf{RQ1.} How well do student LLMs perform, and do they effectively mimic the teacher LLM?

\noindent\textbf{RQ2.} Is augmentation effective in low-data regimes?

\subsection{Models}
\label{subsec:models}

Our \emph{teacher LLM} is GPT-3.5-Turbo. We explored two  \emph{student LLMs}: (1) Llama-2-7b-chat-hf \cite{touvron2023llama2} with Low Rank Adaptations (LoRA) \cite{hu2021lora} and 4-bit quantization. Our Llama-2-7b-chat-hf SFT runs on an NVIDIA Tesla A100 GPU. (2) FlanT5-base (250M) \cite{chung2022scaling}, which is faster and works on an NVIDIA Tesla L4 GPU, which is five times cheaper. We fine-tuned both student LLMs for 10 epochs, selecting the best checkpoint from the last 3 epochs based on the highest final average tool score. See Appendix~\ref{subsec:appendix_student_LLMs_implementation_details} for details.


\subsection{Results}
\label{subsec:results}

\begin{table*}[!t]
    \centering
    \footnotesize
    \begin{tabular}{c|c|c|c|c|c|c}
        \toprule
        \textbf{Row} & \textbf{Model} & \textbf{Test} & \textbf{Adjust} & \textbf{Selective Adjust} & \textbf{Filter} & \textbf{Overall} \\
        \midrule
        1 & \multirow{3}{*}{Llama-2-7b-chat-hf} & All & (\textcolor{gray}{.95}, \textcolor{gray}{.63}, .76) & (\textcolor{gray}{.75}, \textcolor{gray}{.66}, .70) & (\textcolor{gray}{.81}, \textcolor{gray}{.71}, .76) & .74 \\
        2 & & $r_3$ & (\textcolor{gray}{.98}, \textcolor{gray}{.68}, .80) & (\textcolor{gray}{.82}, \textcolor{gray}{.67}, .74) & (\textcolor{gray}{.92}, \textcolor{gray}{.73}, .81) & .78 \\
        3 & & $r_5$ & (\textcolor{gray}{.98}, \textcolor{gray}{.75}, .85) & (\textcolor{gray}{.87}, \textcolor{gray}{.71}, .78) & (\textcolor{gray}{.91}, \textcolor{gray}{.83}, .87) & .83 \\
        \midrule
        4 & \multirow{3}{*}{FlanT5-base (250M)} & All & (\textcolor{gray}{.95}, \textcolor{gray}{.57}, .72) & (\textcolor{gray}{.76}, \textcolor{gray}{.65}, .70) & (\textcolor{gray}{.78}, \textcolor{gray}{.71}, .74) & .72 \\
        5 & & $r_3$ & (\textcolor{gray}{.99}, \textcolor{gray}{.61}, .76) & (\textcolor{gray}{.87}, \textcolor{gray}{.66}, .75) & (\textcolor{gray}{.88}, \textcolor{gray}{.72}, .79) & .77 \\
        6 & & $r_5$ & (\textcolor{gray}{.99}, \textcolor{gray}{.68}, .80) & (\textcolor{gray}{.90}, \textcolor{gray}{.71}, .79) & (\textcolor{gray}{.89}, \textcolor{gray}{.82}, .85) & .81 \\
        \midrule
    \end{tabular}
    \caption{Offline evaluation results for our student models. Metrics include (\textcolor{gray}{tool-selection score}, \textcolor{gray}{quality score}, final score), and the average final score across the tools (\textbf{Overall}). Results show that FlanT5-base performs very similarly to Llama-2-7b-chat-hf, with only a 0.02 gap \textbf{(rows 1, 4)}. Interestingly, both models perform better on a test subset with more popular user intents ($r_5$ > $r_3$ > All), where $r_i$ denotes user intents with at least $i$ calls.}
    \label{tab:eval_results}
\end{table*}

\noindent\textbf{RQ1 (Performance).} 
We begin evaluating our student LLMs on the \emph{test set} with our offline evaluation  (Section \ref{subsec:offline_evaluation}). 
We report results using our metrics (tool-selection score, quality score, final score) per tool in Table~\ref{tab:eval_results}, as well as the overall average final score. We can see both student models achieve comparable performance, despite FlanT5-base being smaller (rows 1, 4). 

We denote by $r_i$ unique user intents with at least $i$ calls. Interestingly, both models perform better on subsets of the test including more popular intents ($r_5$ > $r_3$ > All), This is important for production, as these intents cover more traffic.

Next, we conducted a reality check on a sample of 15 generated images (See Figure~\ref{fig:teacher_examples_for_generated_images_custom_source_images} and Appendix~\ref{subsec:appendix_examples_for_generated_images}). Three calibrated team annotators reviewed each sample according to two criteria: (1) is the image relevant to the intent, and (2) does the student model correctly mimic the teacher. After aggregating the majority vote, we got: Relevance of Teacher: 86.7\%, Llama-2-7b-chat: 86.7\%,  FlanT5-base: 93.3\%. Both students successfully mimicked the teacher 73.3\% times (11 images each, but not the same). These results match Table~\ref{tab:eval_results}, showing our student LLMs have similar performance. 

The average latency for running all tools was 1.63s for Llama-2-7b-chat-hf on an A100 GPU and 1.38s for FlanT5-base on an L4 GPU, both significantly faster than GPT-3.5-Turbo.

\noindent\textbf{A/B tests.}
In addition to offline evaluation, we conducted two online A/B tests. First, we compared our teacher, GPT-3.5-Turbo (tested on 94,317 projects), with Llama-2-7b-chat-hf (93,495 projects). We measured project completion rates as an indicator of user satisfaction\footnote{Satisfaction might be affected by other factors, such as latency.} (Section \ref{subsec:online_evaluation}).
The completion rate for the teacher was 96.1\% of that of Llama-2-7b-chat-hf (no statistical significance). Thus, we conclude they are comparable.


In our second A/B test, we compared our student models. FlanT5-base (tested on 20,294 projects) achieved a completion rate of 99\% of that of Llama-2-7b-chat-hf (20,282 projects). 
Thus, we conclude they are comparable and choose FlanT5-base for its lower latency and cost. Importantly, we are encouraged by the fact that our offline metrics align with the results of the online A/B tests \footnote{Note that the two A/B tests conducted at different times and on partial traffic.}.



\begin{table}[!t]
    \centering
    \footnotesize
    \begin{tabular}{|c|c|c|c|}
        \hline
        \textbf{Train \%} & \textbf{Augmentations} & \textbf{Train Size} & \textbf{Overall} \\
        \hline
        100 & 0 & 8252 & \textbf{.72} \\
        \hline
        \multirow{2}{*}{50} & 0 & 4126 & .68 \\
        & 781 (15.9\%) & 4907 & \textbf{.70} \\
        \hline
        \multirow{2}{*}{25} & 0 & 2063 & .61 \\
        & 784 (27.5\%) & 2847 & \textbf{.66} \\
        \hline
        \multirow{2}{*}{12.5} & 0 & 1031 & .52 \\
        & 806 (43.8\%) & 1837 & \textbf{.65} \\
        \hline
    \end{tabular}
    \caption{FlanT5-base's performance in subsets of the train set, with and without augmentation. We can see that augmentation is effective in limited data increasing the overall score by 0.13 for the 1/8 sample. With larger training subsets, the proportion of augmentations (\%) decreases, reducing overall improvement as expected. }
    \label{tab:augmentation_results}
\end{table}

\noindent\textbf{RQ2 (Augmentation).}
We evaluated FlanT5-base student LLM's performance on different sizes of random training samples using offline evaluation metrics (Section \ref{subsec:offline_evaluation}) and assessed the impact of our data augmentation for each sample (Section \ref{subsec:data_augmentation}). 
Table~\ref{tab:augmentation_results} shows that 
augmentation is highly effective with limited data, increasing the overall score by 0.13 (25\%) for the 1/8 sample. With larger training subsets, the proportion of augmentations decreases, which in turn reduces the overall improvement.

\section{Related Work}

\noindent\textbf{Pre-LLM Dialogue-Based Image Editing.}
Natural language instructions for image editing have been explored extensively, particularly through dialogue systems, prior to the advent of LLMs. For instance, \citet{lin2020adjusting} introduced NLIE, a system designed to convert high-level user commands into precise edits, aiding tasks such as object segmentation and action mapping. Similarly, \citet{kim2022caise} developed Caise, a conversational agent that integrates image search and editing via natural language dialogue. Despite these advancements, both systems struggled with ambiguous or complex instructions and found it difficult to support detailed artistic edits or fully capture user preferences through language alone. 

\noindent\textbf{LLM-Based Tool Invocation for Multimedia Tasks.}
Diffusion-based models for instruction-guided video editing still lag behind image models in visual quality and user control \cite{geyer2023tokenflow, couairon2023videdit, qi2023fatezero, ceylan2023pix2video, kara2024rave}. 
To address this, we drew inspiration from previous research \cite{liu2023llava,wang2023voyager,schick2024toolformer} that used LLMs to invoke tools for complex general and multimedia tasks beyond the LLM's capabilities. The strength of this approach is the LLM's ability to perform diverse visual tasks using tools, which can be integrated into an AI agent at a low development cost.

Two main approaches exist for using tools with LLM planners: (1) tool chaining via prompt engineering and in-context learning \cite{wu2023visual, yang2023mm, caciularu2024tact}, and (2) instruction tuning of LLMs \cite{yang2024gpt4tools, patil2023gorilla, lian2024recai}. Similar to \cite{patil2023gorilla, eldan2023tinystories}, we used a strong LLM proficient with tools through prompt engineering and in-context learning as a teacher to create an instruction tuning dataset for smaller open-source models. A distinctive feature of our approach is incorporating users' behavioral signals in the tuning process.


\label{sec:related_work}

\section{Conclusions and Future Work}

We introduced a novel NLP application for automatic video editing using LLMs, focusing on tonal color adjustment. We fine-tuned a (smaller) student LLM  with guidance from a (larger) teacher LLM, while leveraging user behavioral signals. We proposed offline evaluation metrics and showed that  
our student models succeeded in matching the performance of our teacher model (GPT-3.5-Turbo)
in both offline and online experiments. Our solution significantly reduces costs and latency, crucial for real-time industry applications.

In the future, we plan to test potential fine-tuning improvements by adding rationale as an additional label for supplementary supervision in a multi-task framework, as in \citet{hsieh2023distilling}. We also aim to quantify the benefits of integrating user signals versus relying solely on unfiltered teacher LLM outputs, and to explore other methods for combining user feedback, including {personalization}. We also plan to extend our one-hop responses to conversational agent / dialogue system.
More broadly, we aim to apply our research to additional tools, features, and applications (e.g., light effects, transition between clips, etc.).
Our code and data can be found at our project website: \url{ https://www.orensultan.com/ai_recolor.github.io/}.
We hope to inspire researchers to adopt our best practices in developing novel multi-modal real-time applications using tool chaining.

\paragraph{Acknowledgements}

We would like to express our gratitude to the various teams and stakeholders at Lightricks. Special thanks go to the Machine Learning Infra team for their invaluable support in setting up the online A/B test experiments and enabling production deployment, with particular appreciation to Itay Verkh, Ram Janovski, and Yardena Meymann. We also extend our thanks to the development team, including Guy Niv, Yossy Carmeli, Shoshana Gross, and Dana Fleischer. Additionally, we are grateful to the Product team, especially Rachel HaCohen, Amit Benski, and Alon Yaar. Lastly, we thank the anonymous reviewers for their constructive feedback.

\section*{Ethical Considerations}

Our dataset includes only user intent and model responses. We did not collect any asset  used by users or any personally identifiable information.
In offline evaluation of the images, we used images from the internet or other sources, which were not taken by our users.

\section*{Limitations}

\begin{compactitem}
    \item \textbf{Isolated tools with fixed sequential order}: The current framework employs the three tools independently,  without integrated reasoning, which affects the cohesiveness and effectiveness of the editing process. The tools are applied in a fixed sequence (Adjust, Selective Adjust, LUT filters), which may not be optimal for all scenarios. Training the LLM also to consider dependencies between the tools could improve its flexibility.

    \item \textbf{Overly strict offline metric}: We use cosine similarity to a single ground-truth solution when comparing predicted parameters for the Adjust and Selective Adjust tools, even though multiple different combinations might fulfill the user's request.

    \item \textbf{One-hop responses}: Our current implementation supports one-hop responses, where the user provides a stylistic request in natural language and receives an immediate response. Expanding this to conversational agents or dialogue systems could better adapt to the user's specific needs.

    \item \textbf{Language}: Our dataset contains mostly English user intents. We acknowledge that results may differ in other languages. 

\end{compactitem}

\bibliography{anthology,custom}

\appendix

\section{Appendix}

\subsection{Tools Parameters}
\label{subsec:appendix_tools_parameters}
Our AI Video Filter task is in the domain of tonal color adjustment. Within this scope, our application features three tonal adjustment tools: \emph{global adjust} (global color range), \emph{selective adjust} (selective color ranges), and \emph{filters} (LUTs). Each tool has up to a dozen parameters, which only a professional video/photo editor knows how to set correctly.
The \emph{adjust} tool has 14 parameters, including exposure, contrast, brightness, highlights, shadows, saturation, vibrance, tint, temperature, linearOffset,
hue, bloom, sharpen, structure. 
The \emph{selective adjust} tool features 12 parameters for colors red, orange, yellow, green, cyan, blue, each with saturation and luminance parameters.
The \emph{filter} tool includes two parameters: the name of the filter (out of dozens), and its intensity.
See Figures~\ref{tab:teacher_llm_adjust_prompt},  \ref{tab:teacher_llm_selective_adjust_prompt}, and \ref{tab:teacher_llm_filter_prompt} for the prompts used by our teacher LLM, including the instructions and parameters with their possible range of values for the tools.

\subsection{Distribution of Tool Parameter Values}
\label{subsec:appendix_tools_parameters_values_distribution}
\begin{figure*}[t]
\includegraphics[width=0.99\textwidth]{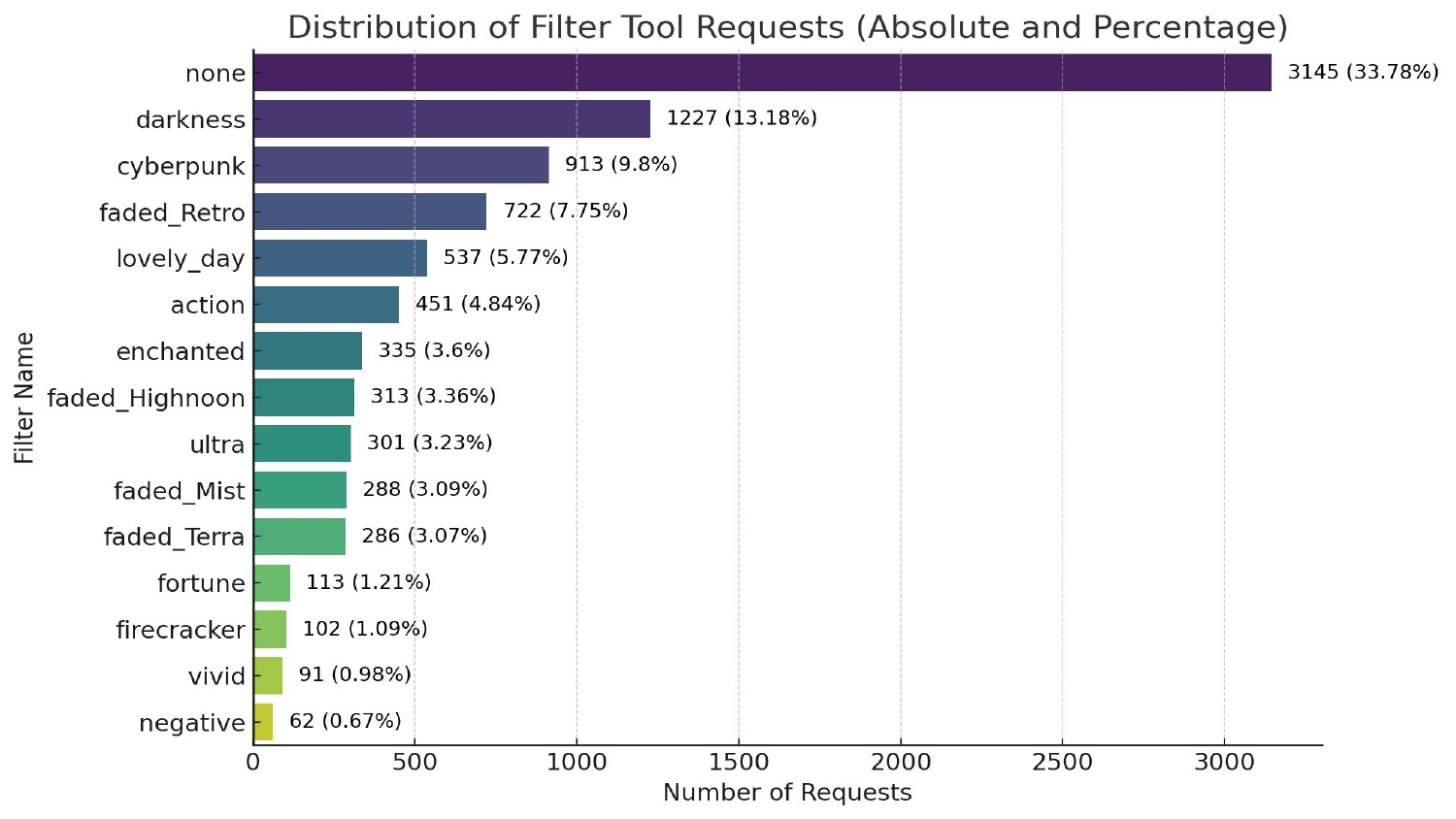}
\centering
\caption{Frequency distribution of the top 15 filter names, ranked by their occurrence in the collected dataset (over 9K instances, each representing a unique user intent along with three tool outputs). Approximately one-third of the cases use the 'none' filter (3,145 instances, accounting for 33.78\%), indicating that the teacher LLM opted not to apply a filter in these instances. Notably, the ``darkness'' and ``cyberpunk'' filters are among the most popular, each accounting for 10\% or more of the user intents. The data also reveals a long-tail distribution.}
\label{fig:filter_dist_of_requests}
\end{figure*}

\begin{figure*}[t]
\includegraphics[width=0.99\textwidth]{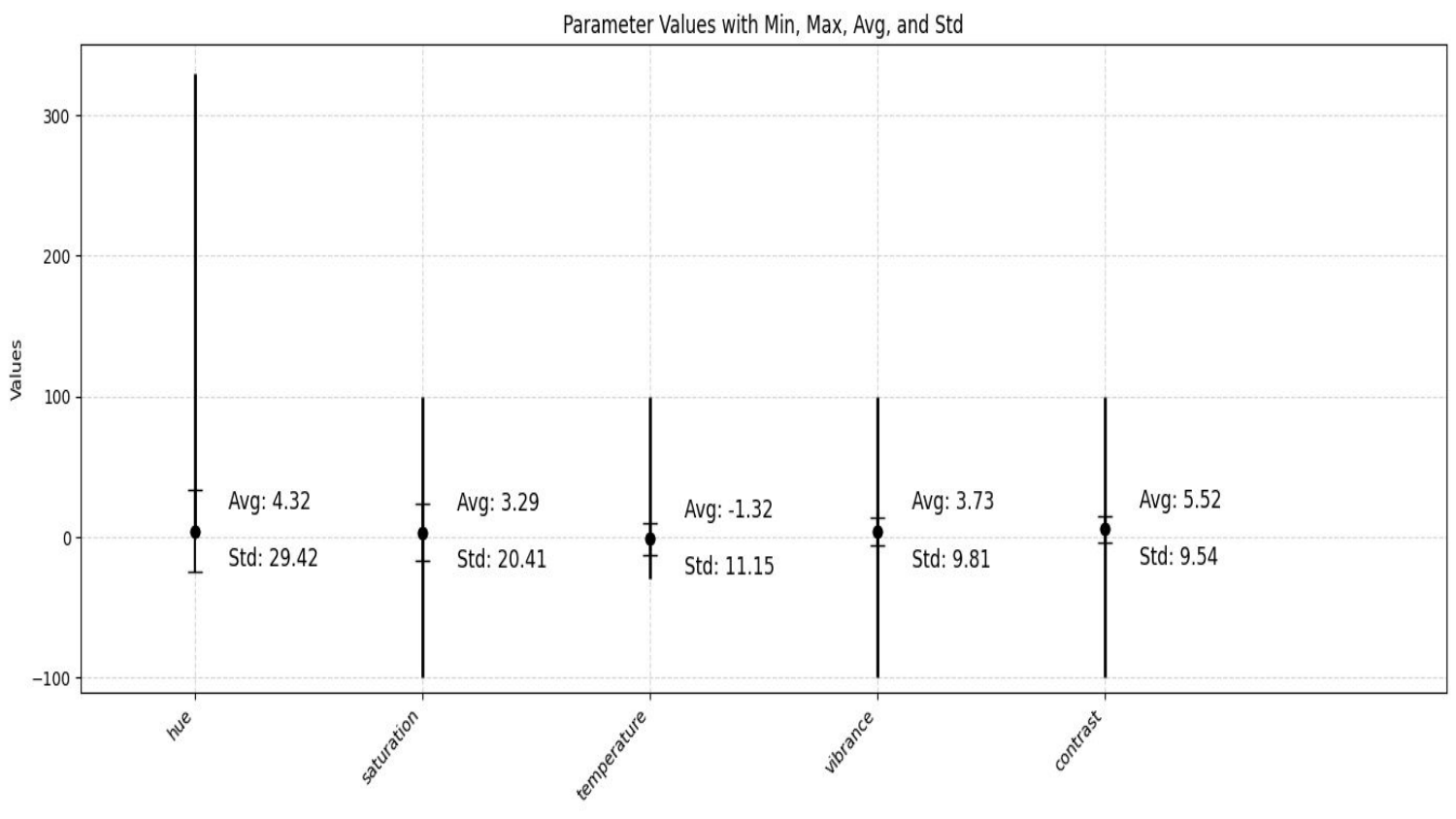}
\centering
\caption{Distribution of values for the adjust tool parameters (top 5 ranked by standard deviation). The vertical black lines indicate the range of values in the dataset, while the dot and inner line represent the average and standard deviation, respectively.}
\label{fig:adjust_dist_of_requests}
\end{figure*}

\begin{figure*}[t]
\includegraphics[width=0.99\textwidth]{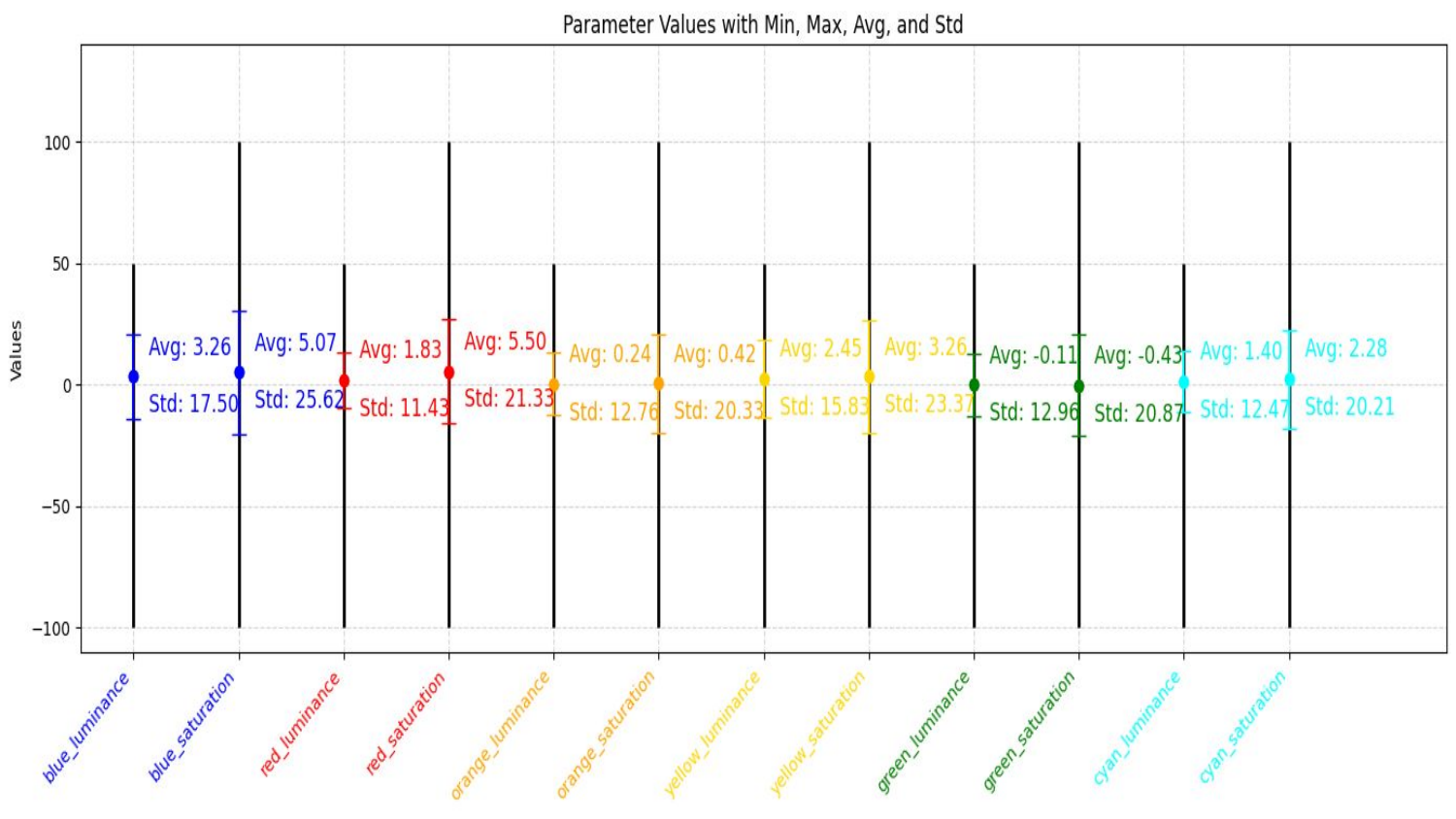}
\centering
\caption{Distribution of values for the selective adjust tool parameters. The vertical black lines indicate the range of values in the dataset, while the dot and inner line represent the average and standard deviation, respectively.}
\label{fig:selective_adjust_dist_of_requests}
\end{figure*}

Figure~\ref{fig:filter_dist_of_requests} illustrates the distribution of filter names observed across the collected dataset for the filter tool.
Additionally, Figures~\ref{fig:adjust_dist_of_requests}, and \ref{fig:selective_adjust_dist_of_requests} present the range, mean, and standard deviation of the parameter values observed across the dataset for the adjust and selective adjust tools. These visualizations emphasize the challenge of parameter prediction, given the extensive variety of filter options and the broad range of continuous parameter values for the adjust and selective adjust tools.

\subsection{Teacher LLM Implementation}
\label{subsec:appendix_teacher_llm_model}
To generate responses (one per tool) for user intents, we used ChatGPT (GPT-3.5-Turbo) as our teacher LLM, with parameters set to temperature=0, max\_tokens=1500, and top\_p=1.

\subsection{Teacher LLM Prompts}
\label{subsec:appendix_teacher_llm_prompts}
See Figures~\ref{tab:teacher_llm_adjust_prompt},  \ref{tab:teacher_llm_selective_adjust_prompt}, and \ref{tab:teacher_llm_filter_prompt} for our teacher LLM (GPT-3.5-Turbo) prompts.


\subsection{Student LLM (Llama-2-7b-chat-hf) Prompts}
\label{subsec:appendix_student_llm_prompts}
See Figures~\ref{tab:student_llm_adjust_prompt},  \ref{tab:student_llm_selective_adjust_prompt}, and \ref{tab:student_llm_filter_prompt} for our Llama-2-7b-chat-hf student LLM prompts.

\subsection{Student LLMs Implementation Details}
\label{subsec:appendix_student_LLMs_implementation_details}
\noindent\textbf{Llama-2-7b-chat-hf.}
For our Llama-2-7b-chat-hf \cite{touvron2023llama2} student LLM, we set the low-rank adaptation dimension to 64, resulting in 33,554,432 trainable params (loraR = 64
loraAlpha = 64, loraDropout = 0.05). We employed 4-bit quantization using the HuggingFace BitsAndBytes (bnb4bitComputeDtype = float16, bnb4bitQuantType = nf4) library to further reduce memory usage. We run the model on NVIDIA Tesla A100 GPU. Important training params are: bf16: false, fp16: true, perDeviceTrainBatchSize: 4, perDeviceEvalBatchSize: 16, gradientAccumulationSteps: 1, maxGradNorm: 0.3, optim: pagedAdamw32bit, learningRate: 4e-5, lrSchedulerType: constant, warmupRatio: 0.03, weightDecay: 0.001, epochs: 10. 

\noindent\textbf{FlanT5-base.}
We run our FlanT5-base (250M) student LLM \cite{chung2022scaling}
on an NVIDIA Tesla L4 GPU, which is five times cheaper. We did not employ LoRA or quantization techniques as this is a much smaller model, and they are not necessary.
Important training parameters are: bf16: false, fp16: false, perDeviceTrainBatchSize: 4, perDeviceEvalBatchSize: 16, gradientAccumulationSteps: 1, maxGradNorm: 0.3, optim: pagedAdamw32bit, learningRate: 4e-5, lrSchedulerType: constant, warmupRatio: 0.03, weightDecay: 0.001, epochs: 10. 

For both models we take the best checkpoint out of the last 3 epochs based on the highest final average score across the tools.

\subsection{Examples of the Generated Images}
\label{subsec:appendix_examples_for_generated_images}
\begin{figure*}[t]
\includegraphics[width=.99\textwidth]{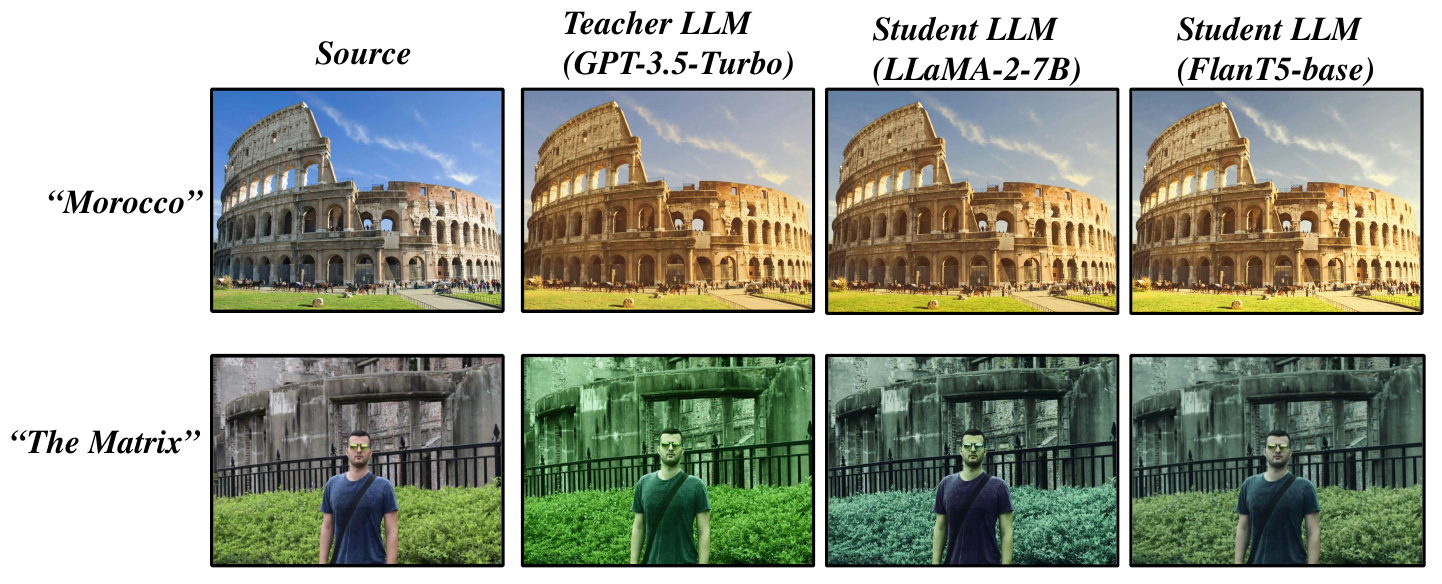}
\centering
\caption{\textbf{Output images for reality check}. Here are examples of samples given to our annotators to evaluate. For each sample, they were asked two binary questions: (1) whether the image is relevant to the intent, and (2) whether the student models correctly mimic the teacher model (see Section~\ref{subsec:results}).
Each sample includes the source image and the outputs of the teacher LLM along with the outputs from both of our student LLMs. Based on the annotator's majority vote: In the first sample: (1) All models produced results relevant to the intent ``Morocco'' (e.g., warm hues, typical of Moroccan landscapes, reflecting its deserts). (2)
Both student models successfully mimicked the teacher LLM.
In the second sample: (1) All models produced results relevant to the intent ``The Matrix'' (e.g., darkness, green tint, and cyberpunk aesthetic) (2) Both student models did not mimic the teacher LLM well.}
\label{fig:teacher_examples_for_generated_images_custom_source_images}
\end{figure*}

See Figure~\ref{fig:teacher_examples_for_generated_images_custom_source_images} for example of samples given to our annotators to check (see Section~\ref{subsec:results}).
Each sample includes the source image and the outputs of the teacher LLM along with the outputs from both of our student LLMs. Based on the annotator's majority vote: In the first sample: (1) All models produced results relevant to the intent ``Morocco'' (e.g., warm hues, typical of Moroccan landscapes, reflecting its deserts). (2)
Both student models successfully mimicked the teacher LLM.
In the second sample: (1) All models produced results relevant to the intent ``The Matrix'' (e.g., darkness, green tint, and cyberpunk aesthetic). (2) Both student models did not mimic the teacher LLM well.

\subsection{GPT-4V Images Evaluation}
\label{subsec:appendix_gpt4v_image_evaluator}
Our goal is to automatically compare two \emph{student LLMs} and determine which one generates parameters that, when applied in our app, produce an image/video that better represents the user's intent.

\begin{figure*}
\centering
\includegraphics[width=.99\linewidth]{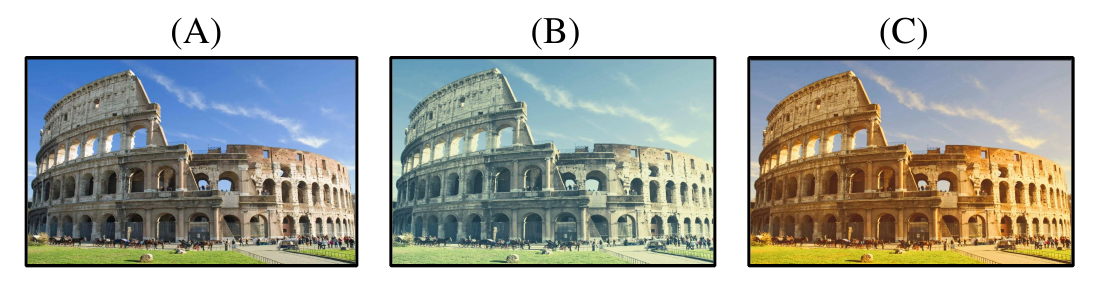}
\resizebox{.99\linewidth}{!}{
\begin{tabular}{p{\linewidth}}  
\toprule
\underline{\texttt{GPT-4V output given the ``Morocco'' intent, and a source image with two edited images.}} \\ \\
\textbf{Prompt:} \\ 
``You are given three images: (A), (B), and (C). 
Describe the transformations of (B) and (C) from (A). 
Then choose which image (B) or (C) better represents the <user's request>'' \\ \\ 
\textbf{Output:} \\ 
\textbf{Image (A)} is the original photo showing the Colosseum in Rome with a clear blue sky and natural daylight colors. \textbf{Image (B)} has undergone a transformation that includes a decrease in contrast, a slight increase in brightness, and a color shift towards a cooler, teal or cyan tint, particularly noticeable in the sky and the shadows. This gives the image a more muted and vintage look. \textbf{Image (C)} has been transformed with an increase in contrast and saturation, giving the colors a warmer, golden hue. This is especially evident in the sky, which has taken on a warm, orange tone, and the overall image has a more vibrant and sunny appearance. \\ \\ 

The Morocco filter look is typically characterized by warm tones, with an emphasis on oranges, reds, and earthy colors that reflect the landscapes and architecture of Morocco. \\ 

\textbf{Answer: (C) better represents the Morocco filter look due to its warm and vibrant color palette.} \\

\bottomrule
\end{tabular}
}
\caption{\textbf{Automatic images evaluation} (see Appendix \ref{subsec:appendix_gpt4v_image_evaluator}). Here is
GPT-4V's output with our Chain-of-Thought prompt, given the user's intent: ``Morocco'' and the three images: (A) Original user's image of the Colosseum. (B) An image generated by applying the first student LLM parameters in our app. (C) An image generated by applying the second student LLM parameters in our app. GPT-4V chose image (C) of the second student LLM, which indeed produced editing that better represents the ``Morocco'' filter look, characterized by a more warm and vibrant color palette like the vibrant colors typical of Moroccan architecture, landscapes, and textiles.}
\label{fig:gpt4v_image_evaluator_example}
\end{figure*}

We initially tried combining different metrics to estimate the aesthetic quality and relevancy of the generated frames, such as the \emph{AestheticScore} \cite{schuhmann2022laion} which predicts people's ratings of images on a scale from 1 to 10, and \emph{PickScore} \cite{kirstain2023pick} which evaluates relevancy based on a preference model trained on text-to-image prompts and user preferences. Ultimately, we chose GPT-4V \cite{openai2023gpt4v} -- a single model which produced us high-quality results.
We asked GPT-4V given the input images A, B, and C (with B and C generated by two different \emph{student LLMs} and A being the original image)
to describe the transformations made for images B and C from image A. Using the Chain-of-Thought (CoT) approach, GPT-4V first described these transformations before determining which image, B or C, better represents the user's intended filter look. See Figure~\ref{fig:gpt4v_image_evaluator_example} for an example. 
Our evaluation metric is simple: we count the number of user intents each \emph{student LLM} wins according to GPT-4V.

\subsection{Mistakes Augmentation LLM Implementation}
\label{subsec:appendix_aug_llm_model}
To generate similar user intents where our student LLM made mistakes, we used GPT4 with a few-shot prompt. The parameters were set to temperature=0, max\_tokens=1500, and top\_p=1.

\subsection{Mistakes Augmentation LLM Prompt}
\label{subsec:appendix_aug_llm_prompt}
See Figure~\ref{tab:aug_llm_prompt} for the few-shot prompt we used to generate the new similar user's intents.


\begin{figure*}
\centering
\resizebox{.99\linewidth}{!}{
\begin{tabular}{p{\linewidth}}  
\toprule
\underline{\texttt{A \textbf{teacher LLM} prompt for the \textbf{global color grading (adjust)} tool.}} \\ 
You are a professional image and video editor. 
Your goal is to make the color adjust filters based on the user's request. The standard tools you can use are: global color grading tool which works globally on all colors, selective color grading (separate adjust for different color ranges) and LUT filter presets. 
Suggest how to use these tools to achieve the requested look.
The tools should be combined and work well when applied together. Write in TOOL whether you should use the color grading tool. If yes, suggest how. \\ 
PARAMETERS: \\ 
"exposure", "contrast", "brightness", "highlights", "shadows", "saturation", "vibrance", "temperature", "tint", "linearOffset" (values between -100 and 100),
"hue" (values between 0 and 360), and "bloom", "sharpen", "structure" (values between 0 and 100).
Is the request asked to apply the look on a specific color range? If so then don't change the "hue" parameter of the global color grading.
If all parameters are zero, then don't use this tool.
If you decide to use the global color grading tool, write in JSON only parameters from PARAMETERS with their correct values.
Otherwise, write an empty json. \\ 
\textbf{Inputs:} USER\_REQUEST \\ 
\textbf{Outputs:} TOOL, JSON \\ 
\\
\textbf{Inputs:} \\ 
\textbf{USER\_REQUEST:} \\ 
a golden hour filter look \\ 
\textbf{Outputs:} \\ 
\textbf{TOOL:} \\ 
Yes, the global color grading tool should be used. \textbf{The temperature should be increased to add a warm, golden tone to the image}. The highlights and shadows should be adjusted to soften the image and mimic the diffused light of the golden hour. The saturation could also be increased to enhance the colors. The hue parameter should not be changed as the request does not specify a specific color range. \\ 
\textbf{JSON:} \\ 
\{ \\
 "exposure": 0, \\ 
 "contrast": 0, \\ 
 "brightness": 0, \\ 
 "highlights": 20, \\ 
 "shadows": -20, \\ 
 "saturation": 10, \\ 
 "vibrance": 0, \\ 
 \textbf{"temperature": 30}, \\ 
 "tint": 0, \\ 
 "hue": 0, \\ 
 "bloom": 0, \\ 
 "sharpen": 0, \\ 
 "structure": 0, \\ 
 "linearOffset": 0 \\ 
\} \\ 
\bottomrule
\end{tabular}
}

\caption{A one-shot, Chain-of-Thought (CoT) prompt for the \textbf{teacher LLM} to generate parameters for the \textbf{global color grading (adjust)} tool. It includes a task description, available tools, and 14 adjustable parameters with specified ranges. The prompt provides an example of a user request for ``golden hour'' with rationale (TOOL) and output parameters (JSON). An empty JSON means the LLM chose not to use the tool. 
We can see that the actions in the reasoning (TOOL) match the parameters (JSON) (e.g., \textbf{``The temperature should be increased to add a warm, golden tone to the image''}). }
\label{tab:teacher_llm_adjust_prompt}
\end{figure*}

\begin{figure*}
\centering
\resizebox{.99\linewidth}{!}{
\begin{tabular}{p{\linewidth}}  
\toprule
\underline{\texttt{A \textbf{teacher LLM} prompt for the \textbf{selective color grading (selective adjust)} tool.}} \\ 
You are a professional image and video editor. Your goal is to make the color adjust filters based on the user's request.
The standard tools you can use are: global color grading tool which works globally on all colors, selective color grading (separate adjust for different color ranges) and LUT filter presets.
Suggest how to use these tools to achieve the requested look.
The tools should be combined and work well when applied together.
The parameters are:
"red", "orange", "yellow", "green", "cyan", "blue". For each we have "saturation" and "luminance" (both between -100 and 100).
Write in TOOL whether you should use the selective color grading tool. If yes, suggest how.
Write in JSON the parameters for selective color grading if you decide to use the selective color grading tool.
You should write the colors ("red", "orange", "yellow", "green", "cyan", "blue"), and a value for each one of the "saturation", "luminance" parameters.
If you decide to use the global color grading tool or LUT filter presets instead, write an empty json. \\ 
\textbf{Inputs:} USER\_REQUEST \\ 
\textbf{Outputs:} TOOL, JSON \\ 
\\
\textbf{Inputs:} \\ 
\textbf{USER\_REQUEST:} \\ 
a golden hour filter look \\ 
\textbf{Outputs:} \\ 
\textbf{TOOL:} \\ 
Yes, the selective color grading tool should be used. We would enhance the reds, oranges, and yellows to create the warm golden hour look.\textbf{ We would also slightly reduce the saturation and luminance of the blues and greens to make the warm colors stand out more.} \\ 
\textbf{JSON:} \\ 
\{ \\ 
  "red": \{ \\ 
    "saturation": 20, \\ 
    "luminance": 10 \\ 
  \}, \\ 
  "orange": \{ \\ 
    "saturation": 30, \\ 
    "luminance": 20 \\ 
  \}, \\ 
  "yellow": \{ \\ 
    "saturation": 40, \\ 
    "luminance": 30 \\ 
  \}, \\ 
  \textbf{"green":} \{ \\ 
    \textbf{"saturation": -20}, \\ 
    \textbf{"luminance": -10} \\ 
  \}, \\ 
  "cyan": \{ \\ 
    "saturation": -20, \\ 
    "luminance": -10 \\ 
  \}, \\ 
  \textbf{"blue":} \{ \\ 
    \textbf{"saturation": -30}, \\ 
    \textbf{"luminance": -20} \\ 
  \} \\ 
\} \\ 
\bottomrule
\end{tabular}
}
\caption{A one-shot, Chain-of-Thought (CoT) prompt for the \textbf{teacher LLM} to generate parameters for the \textbf{selective color grading (selective adjust)} tool. It includes a task description, available tools, and parameters (six colors with two adjustable parameters each, from -100 to 100). The prompt shows an example user request for ``golden hour'' with rationale (TOOL) and output parameters (JSON). An empty JSON means the LLM chose not to use the tool. We can see that the actions in the reasoning (TOOL) match the parameters (JSON) (e.g., \textbf{``We would also slightly reduce the saturation and luminance of the blues and greens...''}).}
\label{tab:teacher_llm_selective_adjust_prompt}
\end{figure*}

\begin{figure*}
\centering
\resizebox{.99\linewidth}{!}{
\begin{tabular}{p{\linewidth}}  
\toprule
\underline{\texttt{A \textbf{teacher LLM} prompt for the \textbf{LUT filter presets (filter)} tool.}} \\ 
You are a professional image and video editor. Your goal is to make the color adjust filters based on the user's request.
The standard tools you can use are: global color grading tool which works globally on all colors, selective color grading (separate adjust for different color ranges) and LUT filter presets.
Available LUT presets are: "none", "lovely\_day", "action", "vivid", "north", "purple\_rain",
"winter", "faded\_Retro", "faded\_HighNoon",  "faded\_Mist", "faded\_Terra", "faded\_Vista",
"faded\_C1", "faded\_AL2", "teal\_and\_orange\_1", "teal\_and\_orange\_1"
"teal\_and\_orange\_2",
"teal\_and\_orange\_3",
"teal\_and\_orange\_4",
"teal\_and\_orange\_5",
"teal\_and\_orange\_6",
"fortune", "duotone\_red", "spring", "duotone\_pink",
"enchanted", "duotone\_green", "ultra", "duotone\_yellow", "firecracker", "duotone\_orange", "cyberpunk", "darkness", "night\_vision", "negative".
Suggest how to use these tools to achieve the requested look.
The tools should be combined and work well when applied together.
Write in TOOL whether the LUT preset is suitable for the requested look.
The default preset is "none". Only use other than "none" LUTs if you find a very close or exact preset. Otherwise use "none" preset.
If duotone was not explicitly mentioned in the request, but you still choose to use a duotone filter, then use very low intensity of 15.
Suggest a specific preset from the list. Try to find a preset name that is semantically close to the requested look. The closer the preset, the more intense it should be.
If the semantic match is quite close, but not exact use intensity of around 40 to 60.
For example for "cool morning" you can choose the presets named "north" or "winter"
because both associated with cold, but the intensity should be low, about 40 to 60 because cool is not exactly north or winter.
But if the request contains the word "winter" then you should choose the winter LUT with high intensity of around 80.
Write in JSON the correct LUT filter preset from the available LUT presets, and the correct intensity between 0 and 100. \\ 
\textbf{Inputs:} USER\_REQUEST \\ 
\textbf{Outputs:} TOOL, JSON \\ 
\\
\textbf{Inputs:} \\ 
\textbf{USER\_REQUEST:} \\ 
a welding mask filter look \\ 
\textbf{Outputs:} \\ 
\textbf{TOOL:} \\ 
\textbf{The "night\_vision" LUT preset seems the most appropriate since night vision also tends to produce greenish images.} \\ 
\textbf{JSON:} \\ 
\{ \\ 
  \textbf{"name": "night\_vision"}, \\ 
  "intensity": 60 \\ 
\} \\ 
\bottomrule
\end{tabular}
}
\caption{A one-shot, Chain-of-Thought (CoT) prompt for the \textbf{teacher LLM} to generate parameters for the \textbf{filter} tool. It includes a task description, available tools, and parameters for the filter tool (filter name from LUT presets and intensity from 0 to 100). The prompt provides a user request example for ``welding mask'' with rationale (TOOL) and output parameters (JSON). Selecting ``none'' as the filter name indicates the LLM decided not to use the tool. As we can see, the reasoning (TOOL) aligns with the parameters (JSON) (\textbf{The "night\_vision" LUT preset seems the most appropriate}).}
\label{tab:teacher_llm_filter_prompt}
\end{figure*}

\begin{figure*}
\centering
\resizebox{.99\linewidth}{!}{
\begin{tabular}{p{\linewidth}}  
\toprule
\underline{\texttt{A \textbf{student LLM} (Llama-2-7b-chat-hf) prompt for the \textbf{global color grading (adjust)}} tool.} \\ 
<s>[INST] You are a professional image and video editor. Your goal is to make the color adjust filters based on the users request.  
The standard tools you can use are: global color grading tool which works globally on all colors, selective color grading (separate adjust for different color ranges) and LUT filter presets.
The tools should be combined and work well when applied together.
The list of parameters for global color grading is: "exposure", "contrast", "brightness", "highlights", "shadows", "saturation", "vibrance", "temperature", "tint",  "linearOffset" (values between -100 and 100), "hue" (values between 0 and 360), "bloom", "sharpen", "structure" (values between 0 and 100). \\ 
\textbf{The user request is:} <user\_request>. \\ 
Your task is to find the correct values for the parameters in order to achieve the user's request.
Is the request asked to apply the look on a specific color range? If so then don't change the "hue" parameter of the global color grading.
If all parameters are zero, then don't use this tool.
If you decide to use the global color grading tool, write "Parameters:" with the name of the parameters and their correct values.
Otherwise, write an empty string.  [/INST] \\
\bottomrule
\end{tabular}
}
\caption{The prompt for a sample of the \textbf{student LLM} (Llama-2-7b-chat-hf) for the \textbf{global color grading (adjust)} tool. It includes a task description, available tools, and the parameters with their optional values for the adjust tool (14 parameters with specified ranges). It also includes a user request (which varies for each sample) and details about writing the output parameters, specifically, writing the values for each of the 14 parameters in a JSON format.}
\label{tab:student_llm_adjust_prompt}
\end{figure*}

\begin{figure*}
\centering
\resizebox{.99\linewidth}{!}{
\begin{tabular}{p{\linewidth}}  
\toprule
\underline{\texttt{A \textbf{student LLM} (Llama-2-7b-chat-hf) prompt for the \textbf{selective color grading}} tool.} \\ 
<s>[INST] You are a professional image and video editor. Your goal is to make the color adjust filters based on the users request.  
The standard tools you can use are: global color grading tool which works globally on all colors, selective color grading (separate adjust for different color ranges) and LUT filter presets.
The tools should be combined and work well when applied together.
The list of parameters for the selective color grading is: "red", "orange", "yellow", "green", "cyan", "blue". For each we have "saturation" and "luminance" (between -100 and 100). \\ 
\textbf{The user request is:} <user\_request>. \\ 
Your task is to find the correct values for the parameters in order to achieve the user's request.
If you decide to use the selective color grading tool, write "Parameters:" with the colors ("red", "orange", "yellow", "green", "cyan", "blue"), and a value for each one of the "saturation", "luminance" parameters.
Otherwise, write an empty string. [/INST] \\ 
\bottomrule
\end{tabular}
}
\caption{The prompt for a sample of the \textbf{student LLM} (Llama-2-7b-chat-hf) for the \textbf{selective color grading (selective adjust)} tool. It includes a task description, available tools, and the parameters with their optional values for the selective adjust tool (six colors with two parameters each, ranging from -100 to 100). It also includes a user request (which varies for each sample) and details about writing the output parameters, specifically, writing the ``saturation'' and ``luminance'' for each of the six colors in a JSON format.}
\label{tab:student_llm_selective_adjust_prompt}
\end{figure*}

\begin{figure*}
\centering
\resizebox{.99\linewidth}{!}{
\begin{tabular}{p{\linewidth}}  
\toprule
\underline{\texttt{A \textbf{student LLM (Llama-2-7b-chat-hf)} prompt for the \textbf{LUT filter presets (filter)} tool.}} \\ 
<s>[INST] The list of LUT presets is: "none", "lovely\_day", "action", "vivid", "north", "purple\_rain", "winter", "faded\_Retro", "faded\_HighNoon", "faded\_Mist", "faded\_Terra", "faded\_Vista", "faded\_C1", "faded\_AL2", "teal\_and\_orange\_1", "fortune", "spring", "enchanted", "ultra", "firecracker", "cyberpunk", "darkness", "night\_vision", "negative".    \\ 
\textbf{The user request is:} <user\_request>. \\  
Your task is to identify the LUT preset that is most semantically similar to the user's request. \\    
In addition, choose an intensity from 0 to 100 (higher intensity indicates greater similarity to the request). \\   
If there's no close match, choose 'none'. \\    
Write "Parameters:", then write a json with two attributes "name" for your chosen LUT preset and "intensity" for its intensity. [/INST] \\
\bottomrule
\end{tabular}
}
\caption{The prompt for a sample of the \textbf{student LLM} (Llama-2-7b-chat-hf) for the \textbf{LUT filter presets (filter)} tool. It includes the available names of the filters, \textbf{a user request} (which varies for each sample), and details about the task of writing the output parameters, specifically writing the name of the filter and its intensity, in a json format.}
\label{tab:student_llm_filter_prompt}
\end{figure*}

\begin{figure*}
\centering
\resizebox{.99\linewidth}{!}{
\begin{tabular}{p{\linewidth}}  
\toprule
\underline{\texttt{A 3-shot prompt for generating new similar user's intent}} \\ 
You are given an input user request (INPUT\_USER\_REQUEST) for a filter look vibe of an image or video.
Your task is to write a suggestion (SIMILAR\_USER\_REQUEST) for a user request which is different from INPUT\_USER\_REQUEST, but share many similar characteristics. \\ \\ 

\textbf{Inputs:} INPUT\_USER\_REQUEST \\ 
\textbf{Outputs:} SIMILAR\_USER\_REQUEST \\ \\ 
\textbf{Inputs:} \\ 
\textbf{INPUT\_USER\_REQUEST:} \\ 
cool morning \\ 
\textbf{Outputs:} \\ 
\textbf{SIMILAR\_USER\_REQUEST:} \\ 
cold tone \\ 
\textbf{Inputs:} \\ 
\textbf{INPUT\_USER\_REQUEST:} \\ 
dark atmosphere \\ 
\textbf{Outputs:} \\ 
\textbf{SIMILAR\_USER\_REQUEST:} \\ 
dark night \\ 
\textbf{Inputs:} \\ 
\textbf{INPUT\_USER\_REQUEST:} \\ 
vintage film \\ 
\textbf{Outputs:} \\ 
\textbf{SIMILAR\_USER\_REQUEST:} \\ 
retro cinema \\
\bottomrule
\end{tabular}
}
\caption{\textbf{Data augmentation prompt. } A 3-shot prompt for our mistakes augmentation LLM (GPT-4). The input is a user's intent that our student LLM made a mistake on (according to ground truth), the output is a new similar user intent.}
\label{tab:aug_llm_prompt}
\end{figure*}

\label{sec:appendix}

\end{document}